\documentclass{article}
\usepackage{balance}
\usepackage{acra}
\usepackage{tabularx}
\usepackage{mathtools}
\usepackage{graphicx}
\usepackage{algorithmicx}
\usepackage{algpseudocode}
\usepackage{algorithm}
\usepackage{amsmath}
\usepackage{booktabs}
\usepackage{multirow}
\usepackage{color,soul}
\usepackage[super]{nth}

\usepackage{subfigure} 
\usepackage{amsmath}
\usepackage{booktabs}
\usepackage{multirow}
 
\title{Crack Detection Using Enhanced Hierarchical Convolutional Neural Networks}
\author{Q. Zhu, M. D. Phung, Q. P. Ha \\ University of Technology Sydney, Australia \\ 
	\{Qiuchen.Zhu;
	Manhduong.Phung; 
	Quang.Ha\}\@ uts.edu.au}

\begin{document}

\maketitle

\begin{abstract}
	Unmanned aerial vehicles (UAV) are expected to replace human in hazardous tasks of surface inspection due to their flexibility in operating space and capability of collecting high quality visual data. In this study, we propose enhanced hierarchical convolutional neural networks (HCNN) to detect cracks from image data collected by UAVs. Unlike traditional HCNN, here a set of branch networks is utilised to reduce the obscuration in the down-sampling process. Moreover, the feature preserving blocks combine the current and previous terms from the convolutional blocks to provide input to the loss functions. As a result, the weights of resized images can be reduced to  minimise the information loss. 
Experiments on images of different crack datasets have been carried out to demonstrate the effectiveness of proposed HCNN.
\end{abstract}

\section{Introduction}
Surface cracks are an important indicator for structural health status of built infrastructure. Prompt detection and repair for cracks could effectively avoid further damage and potential catastrophic collapse. Traditionally, technical inspection is often conducted by specialists which is costly and difficult to proceed especially in hazardous and unreachable circumstances. With recent development and application of UAVs, vision-based systems have been increasingly used in surveillance and inspection tasks, see e.g., \cite{Sankar2015}, \cite{Phung2017}. Integrating image processing into these vehicles for health monitoring of civil structures requires the development of  effective algorithms for crack detection. 

By observation, a crack is a random curve-like pattern with continuity and visible intensity shift to the surrounding area. In geometrics, the randomness of a curve can be expressed as a varying curvature, whereas the intensity shift presents the contrast between crack patterns and non-crack background. Originally, thresholding techniques are applied to solve the crack detection problem by using intensity information \cite{Oliveira2013}. Those techniques work well in the clear background due to the separation of crack pixels in the histogram distribution. However, it severely mislabels the images with noisy background as the feature of non-crack textures usually presents a similar contrast. Moreover, uneven light conditions in photographing and transforming between the colour and greyscale space also leads to strong interference \cite{Kwok2009}.  

Recently, deep convolutional neural networks (DCNN) have been developed to provide a solution that combines both intensity and geometrical information. This technique works effectively in traditional computer vision problems like semantic segmentation due to the multiple levels of abstraction in identifying images. Such promising results motivate the application of deep learning (DL) for vision-based surface inspection taking advantage of the mathematical similarity between image segmentation and crack detection.

In early DCNN application to crack detection, the networks are a sequential model ending with fully connected layers \cite{Zhang2016}. Such architecture requires a lot of computational units since almost all pixels in the image contribute their weights on the prediction for each individual pixel. Furthermore, the feature abstraction generated from middle convolutional layers does not directly propagate to the update of model parameters because the loss function only includes the blurred output from the final layer. This abstraction weakens the preservation of detailed patterns and thus may affect the accuracy of crack feature extraction. Recently, the emerging hierarchical networks showed the improvement in avoiding degradation caused by the blurry effect \cite{Zou2018}. Thus, they have great potential in applications for surface inspection and structural health monitoring.  
   
In this study, we present a new algorithm using HCNN for crack detection by means of UAV imaging. An enhanced end-to-end framework for the networks is proposed to identify potential cracks from aerial images. Experiments on different datasets \cite{Shi2016,Zhu2018} and the images obtained from our UAVs \cite{Hoang2019} have been conducted to demonstrate the advantages of our proposed algorithm compared to existing crack detection algorithms in the literature.

This paper is organized as follows. Section 2 introduces the architecture of the approach and the development of our new crack detection algorithm. Section 3 presents the experimental results and comparison between the proposed method and state-of-the-art crack detection algorithms. Discussions on the obtained results are presented in Section 4 followed by the paper's conclusion given in Section 5.   

\section{Crack Detection Algorithm}
In this section, we introduce the architecture of the proposed hierarchical convolutional neural networks for crack detection, the computation stream of the loss function and the enhancement in the encoder network for preserving image features. This is expected to improve the learning performance in difference with the networks proposed for the DeepCrack \cite{Zou2018}. 

\subsection{Proposed architecture}
The proposed networks are built based on the pipeline the DeepCrack  which inherits the encoder-decoder framework of Segnet \cite{Badrinarayanan2016}. The sequential network of the encoder has 5 convolutional blocks containing 13 convolutional layers in total. For down sampling, each block, which includes two or three $3\times 3$ convolutional layers in series corresponding respectively to a $5\times 5$ or $7\times 7$ convolutional layer,  is followed by a pooling layer that downscales the image and reserves the values and indices of local maxima.  This queue of convolutional layers is eventually equivalent to a single layer, whose number of parameters can be reduced dramatically. Through each block and the corresponding pooling layer, each feature map of the current scale is created and shrinks to a quarter size of the input. Therefore, the size of the receptive field (RF) in the next convolutional layer increases. Consequently, the crack features captured by the blocks become sparser with the enlargement of RF.

The decoder networks is a reflection of the encoder network in a reverse order with the input of each decoder block being processed by an upsampling layer to recover the size of feature map via referring recorded indices. To reconstruct the resolution of image, the following blocks recover the sparse image generated from the last upsampling. Since the indices from pooling layers are saved and transmitted throughout the whole queue, important information of boundaries on the image is preserved.  
\begin{figure*}[h]
	\centering
	\includegraphics[scale=0.36]{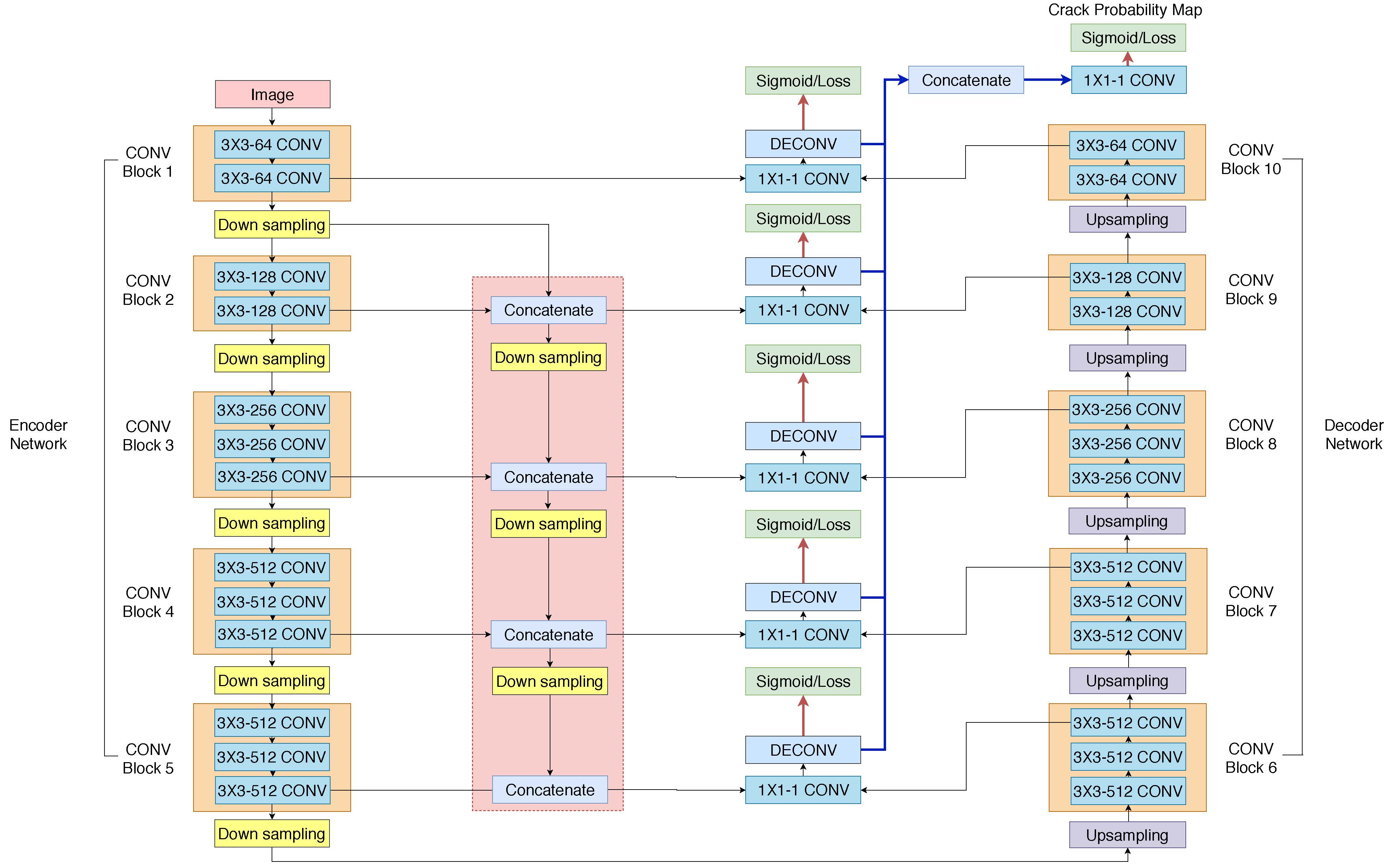}
	\caption{Network architecture}
	\label{fig:net}
\end{figure*} 
To exploit both sparse and detailed feature maps, we propose to set an additional branch in the middle to fuse the outputs from the encoder and decoder blocks. Moreover, the continuous map on the top is directly fed into this branch to augment the low-rank feature map from the encoder to compensate for the feature loss in coarse maps. As shown in Fig.\ref{fig:net}, the downsampled feature map from the upper encoding blocks first concatenates the feature map from the lower hierarchy. The concatenated encoding map and its corresponding decoding map are then compressed into one channel and reshaped to an original-sized feature map for refilling via a $1 \times 1$ convolutional layer and a deconvolutional layer. After this, five original-sized feature maps are integrated through a combination of concatenation and $1 \times 1$ convolutional operations to generate a fusion map $F^k$. Finally, the crack probability map is obtained from the projection of the feature map $F^{fused}$ using a sigmoid function.   

\subsection{Loss function}
As identifying a crack can be considered as a binary segmentation problem containing two classes, crack and non-crack pixels, a binary entropy loss is used to measure the labelling error in the generated crack map. The computation for the entropy loss is conducted in batches. In the training process, one training sample could be expressed as $D = \{(X,Y)\}$, where $X=\left\{x_{i}| i=1, \dots, m\right\}$ represents the pixel values of the original image, $Y=\left\{y_{i}| i=1, \dots, m\right\}$ represents the ground-truth mask of $X$ and $m$ is the number of pixels in one image. For the sake of crack detection, $y_i$ is a binary parameter defined as,
\begin{equation}
y_i = \left\{
\begin{array}{lr}
1, ~ \ x_i  \text{\ is marked as a crack in the mask}\\
0 ~  \text{\ otherwise}.
\end{array}
\right.
\end{equation}

Let $F^k=\{f^k_i|k=1,\dots,5, i=1,\dots,m\}$ and $F^{fused}=\{f^{fused}_i| i=1,\dots,m\}$ be respectively the feature map $f^{k}_i$ at scale $k$ and the fused feature map. The pipeline in Fig. \ref{fig:net} shows the generation of those feature maps. 
The pixel-wise loss as a probability map can be expressed by:
\begin{equation}
l(f_i) =-y_i\log(P(f_i))-(1-y_i)\log(1-P(f_i)),
\label{eq:loss}
\end{equation}
where $P(f_i)$ is the probability of a feature $f_i$ calculated by using the sigmoid function as,
\begin{equation}
P(f_i) = \frac{1}{1+e^{-f_i}}.
\label{eq.sig}
\end{equation} 
Since the labels in the ground-truth data are only 0 and 1, Eq. \ref{eq:loss} can be converted to:
\begin{equation}
l(f_i) = \left\{
\begin{array}{lr}
-\log P(f_{i}), &\   \ y_i=1\\
-\log (1-P(f_{i})), &\   \ y_i=0. 
\end{array}
\right.
\end{equation}
The aim of updating parameters is to train the model so that the output probability maps are close to the ground-truth mask. Therefore, all the probability maps should contribute to the loss function. The overall loss $\mathcal{L}$ of one single image is then obtained from the superposition of pixel-wise loss to every $F^k$ and $F^{fused}$:
\begin{equation}
\mathcal{L}=\sum_{i=1}^{m}\left(l(f_{i}^{fused})+\sum_{k=1}^{5} l(f_{i}^{k})\right).
\end{equation}

\subsection{Enhancement in the encoder network}
The main difference between the proposed networks and the DeepCrack rests with the encoding source for the original-sized feature map. Here, the encoder input is pre-processed in the additional routine block as shown in Figure \ref{fig:net}. On each scale, the encoder output from the upper block iteratively passes the next access in the $1 \times 1$ convolutional merging step with concatenation at the output of the current scale. Therefore, the output from the encoder is half-inherited so that the possession of upper-level features in merging channels increases along with the forward propagation of the convolutional network. To further explain the emphasis on upper-level feature maps, we first discuss the probability model for crack detection in the following.


From the probabilistic perspective, there are two random events, $C_1$ and $C_0$, involved in the crack detection problem, where $C_1$ indicates a crack pixel and $C_0$ implies a non-crack background. Accordingly, two conditional probabilities are defined, the probability $P\left(C_{1}| x_i\right)$ that $x_i$ belongs to a crack and the probability $P\left(C_{0}| x_i\right)$ that $x_i$ belongs to the non-crack background after an observation on pixel $x_i$. They are expressed as: 
\begin{equation}
\begin{aligned}
P\left(C_{1}| x_i\right) & =  \frac{P\left(C_{1}, x_i\right)}{P(x_i)} \\ &=  \frac{P\left(x_i | C_{1}\right) P\left(C_{1}\right)}{P\left(x_i | C_{1}\right) P\left(C_{1}\right)+P\left(x_i | C_{0}\right) P\left(C_{0}\right)}\\&=\frac{1}{1+\frac{P\left(x_i | C_{0}\right) P\left(C_{0}\right)}{P\left(x_i | C_{1}\right) P\left(C_{1}\right)}}\\&=\frac{1}{1+e^{-a(x_i)}},
\end{aligned}
\end{equation}
where 
\begin{equation}
a(x_i) = \operatorname {ln} \frac{P\left(x_i | C_{1}\right) P\left(C_{1}\right)}{P\left(x_i | C_{0}\right) P\left(C_{0}\right)}.
\label{Eq.a}
\end{equation}
Assume that the conditional probabilities follow the Gaussian distribution with the same variance \cite{Murphy:Machine}, we have for $j=0,1$:  
\begin{equation}
\begin{split}
P(x_i|C_j)\sim \mathcal{N}(x_i | \mu_j, \mathbf{\sigma^{2}}) =  \frac{1}{\mathbf{\sigma}\sqrt{2\pi}} \exp \left[-\frac{(x_i-\mu_j)^{2}}{2\mathbf{\sigma}^{2}} \right].
\end{split} 
\label{Eq.gau}
\end{equation}
By substituting Eq. (\ref{Eq.gau}) into Eq. (\ref{Eq.a}), $a(x)$ is solved as follows:
\begin{equation}
\begin{aligned} 
a(x_i) &= \operatorname{ln} P\left(x_i | C_{1}\right)-\operatorname{ln} P\left(x_i | C_{0}\right)+\operatorname{ln} \frac{P\left(C_{1}\right)}{P\left(C_{0}\right)} \\ &= \frac{\left(\mu_{1}-\mu_{0}\right)}{\sigma^{2}} x_i+\frac{\mu_{0}^2-\mu_{1}^{2}}{2\sigma^{2}} +\operatorname{ln} \frac{P\left(C_{1}\right)}{P\left(C_{0}\right)} \\ &=w x_i+w_{0}. \end{aligned}
\label{eq.linear}
\end{equation}  
By comparing Eq. \ref{eq.sig} and Eq. \ref{eq.linear}, we can obtain the expression for features of a crack $f_i$ as:
\begin{equation}
f_i = wx_i + w_{0}, 
\end{equation} 
where $w= \frac{\left(\mu_{1}-\mu_{0}\right)}{\sigma^{2}}$ and $w_{0}=\frac{\mu_{0}^2-\mu_{1}^{2}}{2\sigma^{2}} +\operatorname{ln} \frac{P\left(C_{1}\right)}{P\left(C_{0}\right)}$.\\

Therefore the feature map appears to be linearly-dependent with respect to the input when using the sigmoid function to present the probability map. This is somewhat contradictory to the fact that hidden layers with loss functions represent a non-linear transformation in convolutional networks. To get a moderate solution, it is essential to adequately compensate for the non-linearity before adopting the approach with a linear hypothesis.

Since all hidden convolutional layers are implemented with non-linear activations, the deeper layers' outputs naturally represent highly non-linear relations. As a result, outputs of the deeper encoder networks deviate further from the linear hypothesis, causing a negative impact on the accuracy of pixel-wise predictions. In our proposed model, the enhanced encoder outputs get more weight than the upper-level feature maps in order to reduce nonlinearity. Under the premise of overall non-linearity reduction, this adjustment improves reliability of the probability maps, and as such, resulting in a network model that can approach closer to the required hypothesis.

\section{Experiments}
\subsection{Setup for performance verification}
To verify the effectiveness of the proposed method, a thorough comparison is conducted between our HCNN and a recent deep learning framework for crack detection, the Cracknet-V \cite{zhang2019}, in two datasets. Both methods are trained with the same CrackForest dataset.  
Our implementation is based on Tensorflow \cite{Abadi:tensorflow}, an open source platform for deep learning frameworks. The initialisation method of trainable parameters is "He Normal"\cite{he2015} with initial biases of zeros. The filling method applied to deconvolutional layers is the bilinear interpolation. The training rate for the networks is $10^{-5}$. The learning process is optimised by the stochastic gradient descent method \cite{Johnson2013} with the momentum and weight decay set to 0.9 and 0.0005, respectively. The training is conducted for 20 epochs on an NVIDIA Tesla T4 GPU. This setup is applied to the two methods for comparison. The training time for our proposed one and Cracknet-V is 7 and 9 hours respectively.   

\subsection{Datasets}
Two datasets are used in this study with details given as follows:

\paragraph{CrackForest dataset:}
The dataset \cite{Shi2016} contains 118 crack images of pavements with labelled masks in the size of $600 \times 800$. It is used as the training set and is expanded to 11800 images via data augmentation. For this, we rotate the images with a range from 0 to 90 degrees, flip them vertically and horizontally, and randomly crop the flipped images with a size of $256 \times 256$. 

\paragraph{SYDCrack:}
This dataset contains 170 images of wall and road with cracks collected by our UAVs \cite{Hoang2019}. Due to the safety requirements in flying drones, those images were taken in a safe distance from the surface of the infrastructure surface. As a consequence, the resolution of SYDCrack is lower than CrackForest. The ground-truth masks of SYDCrack were manually marked by two persons. All the images in SYDCrack are used for testing. 

\subsection{Evaluation measures}
Since each test image has the corresponding ground-truth mask, the performance of crack detection is evaluated by a supervised measure, $F$-score \cite{Fawcett2006}. As a commonly-used evaluation measure, the
$F$-score is calculated as,

\begin{equation}
F = 2 \cdot \frac{ {Precision} \cdot  {Recall}}{ {Precision}+ {Recall}},
\end{equation}
where $Precision$ and $Recall$ represent the ratio of correctly-labelled crack pixels among all the predicted crack pixels and the correctly-labelled pixels, respectively. Accordingly, a higher $F$-score indicates a stronger reliability of the segmentation.

Since human-labelled masks may be biased, and thus affecting the quantitative results, an unsupervised measure $Q$-evaluation \cite{Borsotti:Quantitative} is also used to evaluate the performance where the ground-truth image is not required. The $Q$-evaluation for crack segmentation is calculated as, 
\begin{equation}
\begin{aligned} 
& Q(I)=\dfrac{1}{10000(j \times k)} \sqrt{N_c}\\
& \times \sum_{n=1} ^{N_c}\left[\dfrac{e_n^2}{1+\log{A_n}}+\left(\dfrac{N(A_n)}{A_n}\right)^2\right],
\end{aligned} 
\end{equation}
where $I$ is the segmented image; $j\times k$ is the size of the image; $N_c$ is the number of classes in segmentation; $A_n$ is the number of pixels belonging to the $n^{th}$ class; and $N(A_n)$ represents the number of classes that have the same number of pixels as the $n^{th}$ class. With this measure, a smaller $Q(I)$ suggests higher quality of the segmentation result and a better crack detection \cite{Zhu2018}.

\subsection{Results}

Experimental results on the two datasets are presented in the following.

\begin{table*}[h]
	\centering
	\begin{tabular}{@{}l|ll|ll|l@{}}
		\toprule
		\multicolumn{1}{c|}{\multirow{2}{*}{Methods}}& \multicolumn{2}{c|}{F-score} & \multicolumn{2}{c|}{Q-measure}  & \multicolumn{1}{c}{\multirow{2}{*}{\begin{tabular}[c]{@{}c@{}}Training\\ Time\end{tabular}}} \\ \cmidrule(l){2-5}                      & CrackForest & SYDCrack  & CrackForest & SYDCrack& \multicolumn{1}{c}{} \\ \hline
		CrackNet-V &  \multicolumn{1}{c}{0.6127}  &\multicolumn{1}{c|}{0.5605}   &  \multicolumn{1}{c}{2.3679}  & \multicolumn{1}{c|}{2.5080}& \textbf{7 hours}\\
		Proposed   &  \multicolumn{1}{c}{\textbf{0.7807}}  & \multicolumn{1}{c|}{\textbf{0.7393}}   &  \multicolumn{1}{c}{\textbf{2.1901}}  & \multicolumn{1}{c|}{\textbf{2.4588}} & 9 hours\\ \bottomrule
	\end{tabular}
	\label{table:crack}
	\caption{Quantitative results}
\end{table*}

\begin{figure*}[htp]   
	\hspace{50pt}
	\begin{tabular}{cc} 
		\subfigure[]{ 
			\begin{minipage}[t]{0.18\textwidth} 
				\centering 
				\includegraphics[width=1.2in]{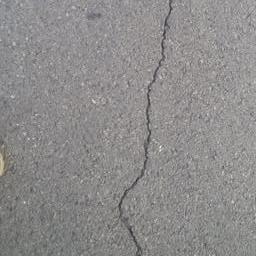}\vspace{2pt}		
				\includegraphics[width=1.2in]{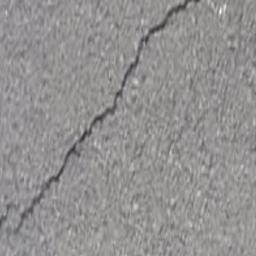}\vspace{2pt}		
				\includegraphics[width=1.2in]{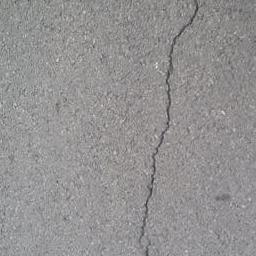}
		\end{minipage}} 
		\subfigure[]{ 
			\begin{minipage}[t]{0.18\textwidth} 
				\centering 
				\includegraphics[width=1.2in]{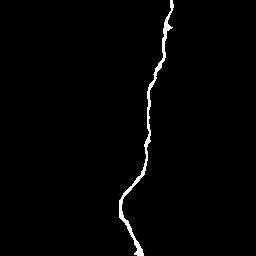}\vspace{2pt}
				\includegraphics[width=1.2in]{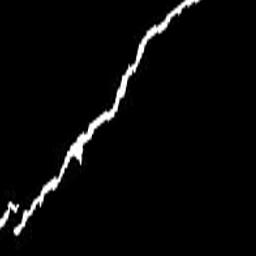}\vspace{2pt}	
				\includegraphics[width=1.2in]{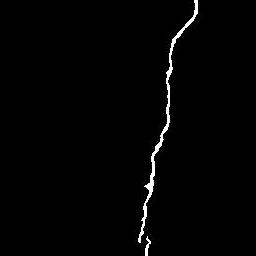} 
		\end{minipage}} 
		\subfigure[]{ 
			\begin{minipage}[t]{0.18\textwidth} 
				\centering 
				\includegraphics[width=1.2in]{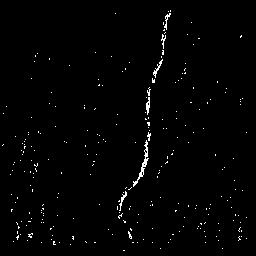}\vspace{2pt}		
				\includegraphics[width=1.2in]{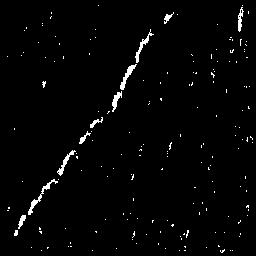}\vspace{2pt}	
				\includegraphics[width=1.2in]{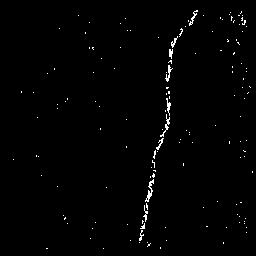}  
		\end{minipage}}
		\subfigure[]{ 
			\begin{minipage}[t]{0.18\textwidth} 
				\centering 
				\includegraphics[width=1.2in]{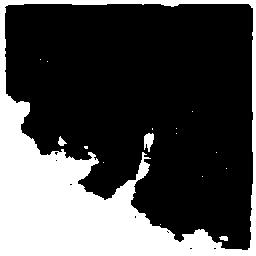}\vspace{2pt}	
				\includegraphics[width=1.2in]{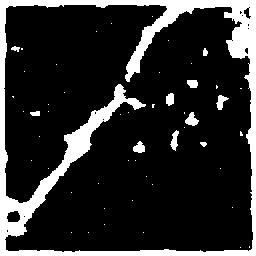}\vspace{2pt}	
				\includegraphics[width=1.2in]{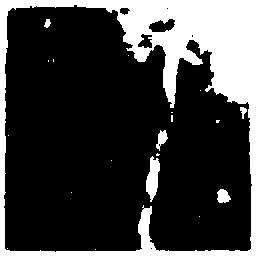} 
		\end{minipage}} 
	\end{tabular} 
	\caption{Crack detection results with CrackForest dataset: (a) original image; (b) ground truth; (c) proposed algorithm; (d) CrackNet-V.}
	\label{fig:crack1}   
\end{figure*}

\paragraph{Results on CrackForest:}
The crack detection results of CrackForest are depicted in Figure \ref{fig:crack1}.
It shows that Cracknet-V is able to extract general crack features but with a bigger width compared to the ground truth. This means neighbourhood pixels were incorrectly labelled as crack. In addition, almost all pixels at the edge of the original image are classified into the crack region. Our proposed method, on the other hand, presents a better matched contour of crack but with some level of isolated noises. Unlike the adjacent noise produced by Cracknet-V, such isolated ones can be easily removed in post-processing.   

\begin{figure*}[htp]  
	\hspace{50pt} 
	\begin{tabular}{cc} 
		\subfigure[]{ 
			\begin{minipage}[t]{0.18\textwidth} 
				\centering 
				\includegraphics[width=1.2in]{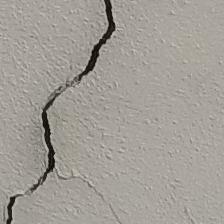}\vspace{2pt}		
				\includegraphics[width=1.2in]{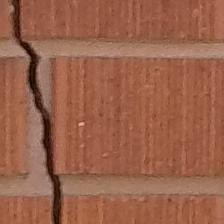}\vspace{2pt}		
				\includegraphics[width=1.2in]{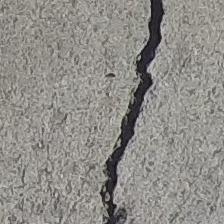}
		\end{minipage}} 
		\subfigure[]{ 
			\begin{minipage}[t]{0.18\textwidth} 
				\centering 
				\includegraphics[width=1.2in]{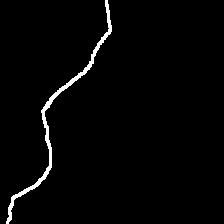}\vspace{2pt}		
				\includegraphics[width=1.2in]{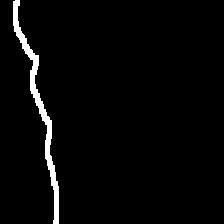}\vspace{2pt}		
				\includegraphics[width=1.2in]{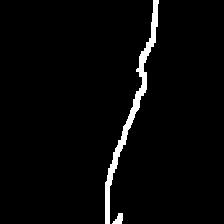}
		\end{minipage}} 
		\subfigure[]{ 
			\begin{minipage}[t]{0.18\textwidth} 
				\centering 
				\includegraphics[width=1.2in]{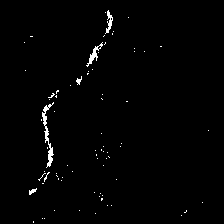}\vspace{2pt}		
				\includegraphics[width=1.2in]{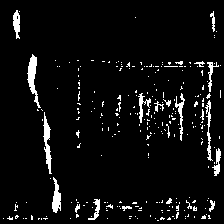}\vspace{2pt}		
				\includegraphics[width=1.2in]{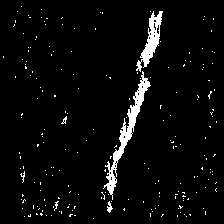}
		\end{minipage}}
		\subfigure[]{ 
			\begin{minipage}[t]{0.18\textwidth} 
				\centering 
				\includegraphics[width=1.2in]{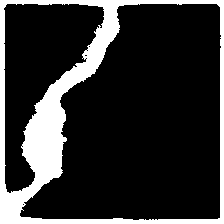}\vspace{2pt}
				\includegraphics[width=1.2in]{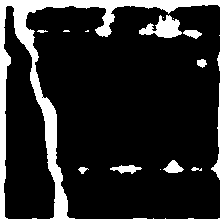}\vspace{2pt}
				\includegraphics[width=1.2in]{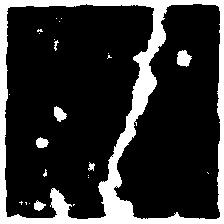}
		\end{minipage}} 	
	\end{tabular} 
	\caption{Crack detection results with SYDCrack dataset: (a) original image; (b) ground truth; (c) proposed algorithm; (d) CrackNet-V.}
	\label{fig:crack2}
\end{figure*}

\paragraph{Results on SYDCrack:}
The detection results of SYDCrack are shown in Figure \ref{fig:crack2}.
It can be seen that both methods are able to extract the main contour of cracks with a certain level of noise. However, it is noted that Cracknet-V's mislabelling on near-crack pixels is more severe with a low resolution of SYDCrack images. The massive amount of false negative samples strongly contributes to a worse $F$-score. Besides, as shown in the second row, although both approaches are strongly interfered by the texture of the brick, our proposed HCNN still keep the noises unadjacent with crack features and thus relax the difficulty in further extraction.

\paragraph{$F$-score and $Q$-measure:}
The $F$-score and $Q$-measure obtained by using the two methods on given test datasets are listed in Table 1. It can be seen that the proposed HCNN obtains a smaller $F$-score and larger $Q$-measure in both datasets. This clearly indicates better performance of our method in terms of accuracy and uniformity. The results also show a lower segmentation quality of both methods on the SYDCrack dataset, which is mainly attributed by the inconsistency in intensity distribution and resolution between the training set and the SYDCrack. Nevertheless, the smaller difference in $F$-score obtained by the proposed method against the Cracknet-V for both datasets in the study implies its advantage in terms of stability and accuracy.

\paragraph{Training time:}
it can be noted that the training time for our model is longer than Cracknet-V as shown in the last column of Table 1. The additional duration is caused by the higher complexity of the proposed networks. \\

\section{Discussion}

\begin{figure*}[htp]  
	\hspace{150pt} 
	\begin{tabular}{cc} 
		\subfigure[]{ 
			\begin{minipage}[t]{0.18\textwidth} 
				\centering 
				\includegraphics[width=1.2in]{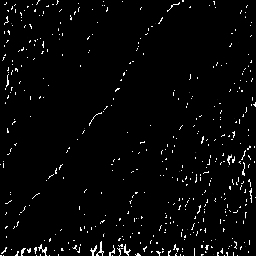}
		\end{minipage}} 	
			\subfigure[]{ 
		\begin{minipage}[t]{0.18\textwidth} 
			\centering 
			\includegraphics[width=1.2in]{image/001_img_44_unet.jpg}
	\end{minipage}} 
	\end{tabular} 
	\caption{results with different training epochs: (a)5 epochs; (b)20 epochs.}
	\label{fig:crack3}
\end{figure*}

Experiment results have indicated that enhanced abstractions from the proposed branch in augmentation to the hierarchical convolutional neural networks (Figure 1) play the main role in improving the accuracy and stability of the proposed method.  Nevertheless, performance of the method is still constrained by the limited epochs available at the demonstration stage. Given more computation power, the number of epochs can be increased to produce a better training model. For the images exemplified in Figure \ref{fig:crack3}, the results with more training epochs have less noise and more well-marked contour. 

Moreover, it can be noticed that performance of the proposed method is affected by scattered noise. The reason is that the generated probability map of our networks is segmented by using a constant threshold of 0.5. That threshold simply divides the crack and non-crack pixels without considering crack clustering. For this, the iterative thresholding method \cite{Zhu2018} can be used to improve it in future research.

Finally, as it can be seen, crack labelling in the ground truth also has a strong influence on the results of crack detection. Further work thus will be to create more accurate crack labels to improve the quality of training data.

\section{Conclusion}
This paper has presented a deep learning framework to identify surface cracks from images collected by UAVs. The enhanced hierarchical convolutional neural networks proposed here can deal with accumulated deviations caused by the non-linearity in deep layers which is the main limitation of existing methods. The key for our improvement is the introduction of a branch network to reduce the non-linear dependency in the deeper convolutional layers. The idea behind this approach is that the upper-layer features are more linear so they should have more weight in labelling. As a result, the proposed approach successfully detected cracks in two datasets from images of different resolutions. The performance is promising in both quantitative and qualitative aspects compared to a benchmark method, the Cracknet-V. This method is promising for potential applications in automatic surface inspection. 

For future work, efforts will be focused on noise removing. For the isolated noise, the clearance can be achieved with size filtering. However, a simple filter may not works for clustered noise like the example shown in Figure 3 (b). In fact, our model is lack of insight in irregular texture since no similar pattern is included in the current training set. In this case, we will extend the training set with more comprehensive information and retrain the network using the pre-trained model. Once a more extracted feature map is obtained, we will attempt to modify the proposed framework to a multitask pipeline that simultaneously accomplish crack detection as well as classification based on severity of the failure.

\section*{Acknowledgements}
The first author would like to acknowledge support from the China Scholarships Council (CSC) for a scholarship and the University of Technology Sydney (UTS) Tech Lab for a Higher Degree Research collaboration grant.
\balance

\end{document}